%% file: main-ioat.tex
\documentclass[11pt,letterpaper]{article}

\usepackage[left=1in,right=1in,top=1in,bottom=0.9in]{geometry}

\usepackage{graphicx}
\usepackage[table,xcdraw]{xcolor}
\usepackage{wrapfig,booktabs}
\usepackage{setspace}
\usepackage{tikz}
\usepackage{amsthm}
\usepackage{tabularx}
\usepackage{longtable}
\usepackage{array}
\usetikzlibrary{decorations.shapes,shapes.geometric,shadows,arrows,automata,positioning,calendar,mindmap,backgrounds,scopes,chains,er,3d,matrix,fit}

\usepackage[utf8]{inputenc}
\usepackage{amsmath}
\usepackage{amsfonts,times}
\usepackage{cite}
\usepackage{lmodern}
\usepackage[most]{tcolorbox}
\usepackage{amssymb}
\usepackage{url}
\usepackage[hidelinks]{hyperref}
\usepackage[font=small,labelfont=bf]{caption}

\usepackage[compact]{titlesec}
    \titlespacing{\section}{0pt}{3ex}{2ex}
    \titlespacing{\subsection}{0pt}{2.0ex}{1.0ex}
    \titlespacing{\subsubsection}{0pt}{1.0ex}{0.5ex}

\graphicspath{{figures/}}

\newcolumntype{L}{>{\centering\arraybackslash}m{3cm}}
\newcolumntype{Y}{>{\raggedright\arraybackslash}X}

\newcommand{\cU}{\mathcal{U}}

\newcommand{\cW}{\mathcal{W}}
\newcommand{\cX}{\mathcal{X}}

\newcommand{\cZ}{\mathcal{Z}}

\newtheorem{theorem}{Theorem}
\newtheorem{definition}[theorem]{Definition}
\newtheorem{proposition}[theorem]{Proposition}

\title{Internet of Agentic Things:\\ Networked AI Agents for Closed-Loop IoT Orchestration}
\author{
Quanyan Zhu\thanks{Department of Electrical and Computer Engineering, New York University, Tandon School of Engineering, Brooklyn, NY 11201, USA. Email: \texttt{qz494@nyu.edu}.}
}
\date{}

\begin{document}

\maketitle
\setstretch{1.55}
\pagenumbering{arabic}

\begin{abstract}
The Internet of Things (IoT) has traditionally been organized as a passive sensing and actuation infrastructure: devices collect data, dashboards display events, and controllers execute predefined logic. The emergence of agentic artificial intelligence creates a different paradigm. In an Internet of Agentic Things (IoAT), networked AI agents perceive the physical environment, reason over user intent and system context, invoke tools and other agents, coordinate distributed IoT APIs, execute actions, and adapt through feedback. This paper develops an article-length framework for IoAT based on the slide concept of Internet of Agentic IoT Networks and positions IoAT as a networked infrastructure for Physical AI at cyber-physical scale. The framework connects three layers: a physical IoT layer containing sensors, actuators, and embedded controllers; an edge/fog layer containing OpenClaw-style orchestration agents, situation-awareness agents, and historians; and a cloud layer containing planning, digital-twin, expertise, and computation agents. We formalize IoAT operation as a coupled workflow-control problem in which user intent is unfolded into agentic subtasks and physical control specifications, while IoT outcomes are folded back into agent states for monitoring, reuse, and replanning. A smart-building after-hours orchestration example illustrates how natural-language intent can become coordinated control of HVAC, lighting, locks, cameras, and alarms. The paper concludes by identifying key research challenges in reliable planning, cross-layer control, safety, trust, privacy, latency, and resilience against adversarial manipulation.
\end{abstract}

\section{Introduction}
\label{sec:introduction}
\input{intro}

\section{Literature Review}
\label{sec:literature}
\input{literature_review}

\section{From Internet of Agents to Internet of Agentic Things}
\label{sec:agents-to-ioat}
\input{internet_agents}

\section{Architectural Framework}
\label{sec:architecture}
\input{architecture}

\section{Smart-Building Orchestration}
\label{sec:usecase}
\input{usecase}

\section{Workflow-Control Model}
\label{sec:model}
\input{workflow_control}

\section{Reliability, Safety, and Trust}
\label{sec:resilience}
\input{resilience}

\section{Research Challenges}
\label{sec:agenda}
\input{research_challenges}

\section{Conclusion}
\label{sec:conclusion}

The Internet of Agentic Things reframes IoT as an active cyber-physical intelligence network rather than a passive device network. The central shift is not merely the addition of a large language model interface. It is the coupling of distributed agentic reasoning with physical sensing, actuation, memory, digital-twin simulation, feedback control, and adaptive replanning. This coupling allows user intent to be translated into executable workflows while preserving the closed-loop discipline required by safety-critical IoT systems. It also gives Physical AI a scalable network form: instead of locating intelligence in a single embodied machine, IoAT distributes embodiment, perception, reasoning, and action across connected devices, edge agents, cloud agents, and human operators.

The resulting architecture is naturally hierarchical. Cloud agents provide reusable strategic intelligence, edge agents provide real-time orchestration and context awareness, and the IoT layer provides physical execution and feedback. The OpenClaw agent plays the role of an agentic controller that mediates among user requests, planning agents, digital twins, historians, and device APIs. In the Physical AI interpretation, OpenClaw is the local coordinator of a distributed body whose limbs are locks, cameras, HVAC systems, lights, robots, sensors, gateways, and actuators. This view makes IoAT a promising direction for smart buildings, energy systems, healthcare facilities, transportation networks, manufacturing plants, and other environments where physical systems must respond intelligently to high-level goals under uncertainty.

At the same time, IoAT creates new technical and governance problems. Agentic workflows can fail through prompt injection, compromised tools, stale memory, inaccurate digital twins, latency, cascading physical effects, and unsafe delegation. These failures are especially consequential because Physical AI turns cognitive errors into physical actions. For this reason, the next generation of IoAT systems must integrate formal planning, model predictive control, runtime monitoring, access control, provenance, anomaly detection, redundancy, and human-aligned oversight. The long-term opportunity is a scalable and trustworthy cyber-physical ecosystem in which every connected device can participate in perception, reasoning, action, and adaptation.

\bibliographystyle{abbrv}
\bibliography{refs}

\end{document}

%% file: intro.tex
The Internet of Things (IoT) has become a foundational infrastructure for modern cyber-physical systems. Cameras, thermostats, smart meters, locks, vehicles, medical devices, industrial controllers, robotic platforms, and environmental sensors continuously sense and influence the physical world. Over the past two decades, IoT has transformed isolated devices into interconnected networks capable of collecting large volumes of operational data and enabling new forms of automation. Yet much of the deployed IoT stack remains organized around a limited operational pattern: devices produce telemetry, data are transported to cloud services or supervisory systems, dashboards provide visibility, rule engines trigger predefined responses, and human operators or fixed controllers decide what happens next. This architecture is useful for monitoring and event reporting, but it remains largely reactive. It is limited as a substrate for autonomous coordination, especially in environments where user goals are high level, device capabilities are heterogeneous, and the physical situation changes continuously.

This limitation becomes more pronounced as cyber-physical environments grow in scale and complexity. Modern smart buildings, hospitals, factories, transportation networks, and energy infrastructures are no longer collections of isolated devices. They are interconnected socio-technical ecosystems whose components interact through physical dynamics, communication networks, software services, organizational policies, and human operators \cite{zhubasar2024sociotechnical,basar2026sociotechnical}. The challenge is no longer merely collecting data. It is transforming distributed sensing into coordinated intelligence that can reason about objectives, anticipate future states, orchestrate actions across many devices, and adapt continuously when conditions change. Traditional IoT architectures were designed primarily to connect things. The next generation of cyber-physical systems must coordinate autonomous decision-making among things, agents, and humans.

The evolution toward IoAT can be understood as a sequence of increasingly sophisticated integrations between computation and the physical world. Early IoT architectures emphasized device connectivity and telemetry collection. Cyber-physical systems added the discipline of feedback between computation and physical dynamics \cite{lee2008cps}. Edge and fog computing responded to latency, bandwidth, privacy, and resilience limits by moving computation closer to devices \cite{bonomi2012fog,shi2016edge,satyanarayanan2017emergence}. Digital twins then introduced synchronized virtual representations of physical assets and processes, making it possible to simulate candidate interventions before acting on the real system \cite{tao2019digitaltwin}. Foundation models and large language models have recently added new capabilities for language understanding, reasoning, planning, tool use, and interaction \cite{bommasani2021opportunities,yao2023react,schick2023toolformer}. In parallel, the idea of Physical AI has emerged to describe AI systems that perceive, reason, learn, and act through real physical embodiment or physical interfaces \cite{li2021physicalai,bousetouane2025physicalaiagents}. The Internet of Agentic Things is the convergence of these developments into an architecture in which autonomous AI agents are embedded across cloud, edge, and device layers and coordinate sensing, reasoning, decision-making, and actuation.

Physical AI and IoAT are closely related, but they emphasize different levels of organization. Physical AI focuses on the capacity of intelligent systems to act in and learn from the physical world, as in robots, autonomous vehicles, surgical systems, smart machines, and embodied assistants. IoAT focuses on the networked infrastructure that allows many such physically grounded agents, devices, digital twins, and human operators to coordinate safely. A single robot can be a Physical AI system; a smart building, hospital, factory, or transportation network becomes an IoAT system when many physical agents and devices are connected through shared perception, planning, governance, and feedback. In this sense, IoAT can be understood as a distributed architecture for scalable Physical AI.

Agentic AI changes the role of intelligence in this architecture. An AI agent is not only a predictor or text generator; it can maintain context, reason over objectives, call external tools, communicate with other agents, execute multi-step plans, monitor outcomes, and revise its behavior based on experience. Multi-agent systems have long studied the design of autonomous interacting entities \cite{wooldridge2009introduction,jennings2000agent}. Recent advances in foundation models, reasoning-and-acting language agents, tool-using models, and LLM-based autonomous agents make such systems easier to instantiate across cloud, edge, and local device environments \cite{bommasani2021opportunities,yao2023react,schick2023toolformer,xi2023rise,zhu2026ioai}. The implication for IoT is significant: instead of treating devices as endpoints connected to dashboards, IoT systems can be embedded in networks of agents that discover capabilities, compose services, coordinate actions, and learn from execution.

We use the term \emph{Internet of Agentic Things} (IoAT) to describe this emerging architecture. IoAT combines the physical reach of IoT \cite{atzori2010iot,gubbi2013iot}, the feedback discipline of cyber-physical systems \cite{lee2008cps}, the locality of edge intelligence, the predictive power of digital twins, the physical grounding of Physical AI, and the task-driven reasoning of agentic AI. In IoAT, agents do not merely observe the physical world; they participate in a closed loop of perception, reasoning, action, and adaptation. The loop begins with sensor data and user intent, passes through agentic planning and tool invocation, reaches physical actuation, and returns through monitoring, outcome evaluation, and replanning. It resembles the observe, orient, decide, and act pattern used to describe adaptive decision-making in complex environments. Sensors and data streams provide observations of the physical world; agents orient those observations relative to goals, policies, and constraints; planning and coordination agents decide among candidate actions; actuators execute the selected actions; and feedback mechanisms evaluate outcomes and trigger adaptation. Unlike traditional control loops that operate primarily on numerical state variables, IoAT agents can reason simultaneously about symbolic goals, contextual knowledge, operational policies, and physical constraints.

A distinguishing feature of IoAT is that perception, cognition, and actuation are distributed across many agents rather than concentrated in a single controller. Each agent may have a perceptual interface to sensors, data streams, histories, or digital twins; a cognitive layer for memory, reasoning, planning, and uncertainty management; and an actuation interface to APIs, controllers, robots, or physical devices. This architecture echoes Physical AI's emphasis on perception, cognition, and action, but extends it from a single embodied system to a networked cyber-physical ecology. These agents may also communicate with one another, forming a distributed ecosystem for collaborative problem solving. A building-energy agent, for example, may consult an occupancy-monitoring agent, negotiate with a facility-management agent, query a digital-twin simulation agent, and coordinate with security agents before changing HVAC operation. In this sense, the IoT network becomes an adaptive decision-making system rather than a passive substrate for data collection.

\subsection*{Motivation}

The motivation for IoAT comes from a mismatch between the complexity of modern IoT environments and the relatively static way many IoT systems are operated. Smart buildings, hospitals, factories, transportation systems, and energy infrastructures are increasingly populated by heterogeneous devices with different vendors, protocols, data formats, failure modes, and safety requirements. A useful decision in such an environment rarely involves a single API call. A request such as ``reduce building energy consumption without affecting occupant comfort'' may require occupancy prediction, weather forecasting, utility-price awareness, HVAC constraints, maintenance schedules, cybersecurity policies, and user preferences. It usually requires interpreting a high-level goal, retrieving local context, checking policies, predicting physical consequences, coordinating multiple devices, monitoring outcomes, and adapting when conditions change. Conventional dashboards and rule engines are not designed to perform this end-to-end reasoning loop.

A second motivation is locality. Many IoT decisions are time sensitive and context dependent. Sending all data to a centralized cloud service can increase latency, bandwidth cost, privacy exposure, and operational fragility. Edge and fog computing were introduced precisely to bring computation, storage, and decision support closer to devices and users \cite{bonomi2012fog,shi2016edge,satyanarayanan2017emergence}. IoAT builds on this architectural idea but adds agentic reasoning. Edge agents do not only preprocess data; they interpret situations, invoke remote expertise when needed, execute local control actions, detect anomalies, and preserve safe fallback behavior when remote services are unavailable. Cloud resources remain important for large-scale optimization, digital-twin simulation, foundation-model services, and long-horizon planning, but many routine or safety-critical operational decisions should be made close to the physical environment.

A third motivation is compositionality and specialization. Future cyber-physical systems require expertise spanning engineering, cybersecurity, operations, maintenance, compliance, economics, and human factors. No single agent is likely to contain all relevant knowledge or authority. A building-energy agent, a security agent, a digital-twin agent, a maintenance agent, and a compliance agent may each be best suited for different parts of a task. IoAT treats these agents as networked services that can be discovered, invoked, and composed. This compositional view mirrors how human organizations manage complex systems and becomes especially important when physical systems cross organizational boundaries, as in campuses, hospitals, industrial supply chains, or city-scale infrastructure \cite{zhu2026ioai,yangzhu2026ioaiworkflow}.

A fourth motivation is predictive orchestration through digital twins. Modern cyber-physical systems increasingly rely on synchronized virtual models that mirror physical assets and processes. Within IoAT, digital twins are not merely passive simulation tools. They become active services that agents can query to evaluate candidate actions, forecast future states, test recovery strategies, and estimate risk before acting on the physical system. This capability is especially valuable in safety-critical infrastructure, where experimental actions may be costly or dangerous. Agentic reasoning coupled with digital-twin simulation allows IoAT systems to test plans virtually, select safer actions, and update memories with both predicted and observed outcomes.

The final motivation is resilience. Agentic control of IoT can create new failure modes, including unsafe tool use, prompt injection, stale memory, digital-twin mismatch, latency-induced instability, adversarial manipulation, coordination failure, and cascading physical effects. These risks are also central to Physical AI, because an error in reasoning can become an unsafe physical action. At the same time, agentic networks can improve resilience by enabling monitoring, explanation, redundancy, adaptive replanning, mutual checking among agents, and human escalation. Multi-agent architectures can reduce single points of failure while enabling adaptive responses to disruptions, consistent with recent work connecting cyber resilience with game, control, learning, and agentic AI frameworks \cite{zhu2024cyberresiliencefoundations,zhubasar2024resiliencerobustness,lidzhu2025agenticcyberresilience}. The central research question is therefore not whether AI agents can be connected to IoT devices, but how they can be connected in a way that is reliable, governable, secure, and physically safe.

The slide concept behind this paper centers on an OpenClaw-style agentic controller operating at the edge of a cyber-physical environment. OpenClaw receives natural-language or application-level intent, discovers relevant capabilities, queries sensing agents and historians, invokes cloud planning agents and digital-twin agents when needed, coordinates IoT APIs, and supervises physical execution through continuous monitoring and feedback. This agent is not a monolithic brain. It is a local orchestrator embedded in a larger A2A network of specialized agents and a physical IoT network of sensors and actuators. It bridges high-level human intent and low-level physical control by translating goals into coordinated workflows spanning both digital and physical domains.

From this perspective, IoAT can be viewed as the natural successor to conventional IoT and as an infrastructure layer for Physical AI. Whereas traditional IoT connects things, IoAT connects autonomous decision-makers. Whereas IoT emphasizes telemetry, IoAT emphasizes closed-loop autonomy. Whereas IoT often treats intelligence as an external service, IoAT embeds intelligence throughout the network. The resulting architecture offers the potential for scalable, adaptive, resilient, and explainable cyber-physical systems capable of operating in environments whose complexity increasingly exceeds the capacity of centralized management approaches.

The objective of this paper is to elaborate that idea into a research article. The contributions are fourfold. First, we articulate the transition from the Internet of AI agents to the Internet of Agentic Things and clarify IoAT's connection to Physical AI. Second, we propose a multi-layer IoAT architecture spanning cloud intelligence, edge/fog orchestration, and physical IoT execution. Third, we formalize IoAT operation as a coupled workflow-control problem in which agentic planning and physical control exchange specifications and feedback. Fourth, we identify reliability, safety, trust, and resilience challenges that must be solved before IoAT can be deployed in high-consequence environments.

The rest of the paper is organized as follows. Section~\ref{sec:literature} reviews the evolution from IoT orchestration and cyber-physical systems toward agentic AI and multi-agent coordination. Section~\ref{sec:agents-to-ioat} explains the move from networked AI agents to agentic IoT. Section~\ref{sec:architecture} presents the layered architecture. Section~\ref{sec:usecase} develops a smart-building orchestration use case. Section~\ref{sec:model} introduces a workflow-control model for IoAT. Section~\ref{sec:resilience} discusses reliability, safety, and trust. Section~\ref{sec:agenda} concludes the technical development with research challenges.

%% file: literature_review.tex
The literature leading to IoAT can be read as four converging lines of work: the evolution of IoT orchestration, the maturation of cyber-physical systems, the rise of Physical AI, and the recent rise of agentic AI and multi-agent coordination. Each line contributes part of the IoAT concept, but none is sufficient on its own. IoT supplies the physical reach, CPS supplies the feedback discipline, Physical AI supplies the embodiment and real-world action perspective, and agentic AI supplies goal-directed reasoning, tool use, and adaptive coordination.

\subsection{IoT Evolution and Orchestration}

The Internet of Things has expanded rapidly over the last decade, with connected devices now spanning consumer electronics, smart homes, industrial automation, transportation, healthcare, energy systems, and critical infrastructure \cite{nageshwaran2026agentic}. Early IoT deployments emphasized connectivity, telemetry, and basic automation. Sensors collected data, gateways transmitted events, dashboards displayed operational status, and predefined controllers or rule engines executed fixed actions. This architecture was effective for visibility and routine automation, but it was not designed for environments in which heterogeneous devices, dynamic objectives, and safety constraints must be coordinated continuously.

As IoT ecosystems have grown in scale and heterogeneity, orchestration has become a central requirement \cite{symphonica2024iot,nageshwaran2026agentic}. IoT orchestration refers to the coordinated management of devices, applications, communication channels, data flows, cloud services, and operational policies across distributed environments. Automated orchestration frameworks can support provisioning, configuration management, data routing, event handling, service composition, and fault response, thereby reducing human intervention and improving reliability \cite{symphonica2024iot}. Such capabilities are particularly important in systems with large-scale data streams and coupled operational dependencies, such as smart grids, industrial control systems, hospital infrastructure, and transportation networks.

Cyber-physical systems provide a closely related foundation. CPS research emphasizes real-time interaction between computational components and physical processes through sensing, communication, computation, actuation, and continuous feedback \cite{lee2008cps,gartner2024cps}. Gartner describes CPS as systems that orchestrate sensing, computation, control, networking, and analytics to interact with the physical world \cite{gartner2024cps}. Modern IoT deployments increasingly exhibit CPS characteristics, especially in autonomous vehicles, robotics, industrial automation, energy systems, and healthcare environments where physical consequences and safety-critical operations are central concerns.

Despite these advances, conventional IoT and CPS architectures often remain tied to centralized logic, static workflows, manually designed policies, and predefined control scripts. They can automate repeated actions, but they generally have limited capacity to interpret high-level intent, reason over context, select among alternative courses of action, and replan when the environment changes. This limitation motivates the integration of autonomous agents into IoT infrastructures.

\subsection{Agentic AI and Multi-Agent Systems}

Recent advances in artificial intelligence have renewed interest in agentic AI, which shifts AI systems from passive inference toward autonomous entities that pursue goals over extended time horizons \cite{zhu2026ioai}. A conventional AI service may answer a prompt or classify an input, but an agentic AI system maintains task context, invokes external tools, queries memory, interacts with other agents, monitors outcomes, and adapts its plan based on feedback. This agentic view is reinforced by recent work on language agents, tool-augmented reasoning, and foundation-model-based autonomous systems \cite{yao2023react,schick2023toolformer,xi2023rise,zhu2026ioai}.

The emergence of agentic AI connects naturally to the long-standing field of multi-agent systems, in which multiple autonomous entities coordinate to achieve individual or collective objectives \cite{wooldridge2009introduction,jennings2000agent}. Contemporary multi-agent architectures differ from many earlier systems because LLMs, retrieval systems, code execution, API access, and tool-use mechanisms make agents more flexible in interpreting tasks, communicating intent, and composing services. IBM characterizes AI agent orchestration as the coordination of specialized AI agents within a unified system so that collective goals can be accomplished efficiently \cite{ibm2025agent}. In such architectures, agents may specialize in planning, perception, execution, optimization, security analysis, explanation, or domain-specific operation.

Recent surveys describe a transition from isolated AI models toward ecosystems of interacting agents \cite{wang2025survey,mdpi2024agentreview}. Zhu's Internet of Agentic AI framework develops this view as a networked environment in which autonomous agents discover one another, negotiate responsibilities, exchange contextual information, invoke tools, and coordinate workflows across distributed infrastructures \cite{zhu2026ioai}. Yang and Zhu further formalize this direction through incentive-compatible distributed teaming, where heterogeneous agents form coalitions under capability, locality, and economic constraints \cite{yangzhu2026ioaiworkflow}. The emphasis is decentralization: intelligence is not concentrated in a single monolithic model, but emerges from the interaction of specialized agents, each with its own data access, capabilities, locality, trust assumptions, and operational constraints.

For IoT, this shift is consequential. IoT environments already contain distributed sensing, actuation, edge gateways, cloud services, enterprise systems, and human operators. Agentic AI provides a way to bind these components into goal-directed workflows. A building agent can ask a weather-risk agent for forecasts, invoke a digital-twin agent to simulate energy strategies, query a security agent for access-control policies, and coordinate device APIs through an edge orchestrator. This is qualitatively different from connecting a chatbot to a device dashboard; it turns the IoT environment into a network of task-capable services.

\subsection{Physical AI and Embodied Cyber-Physical Intelligence}

Physical AI provides a complementary perspective by emphasizing intelligence that is grounded in the physical world. Rather than treating AI as a purely digital service that consumes data and emits text, recommendations, or classifications, Physical AI concerns systems that perceive physical context, reason over embodied constraints, and act through robots, vehicles, machines, controllers, or other physical interfaces \cite{li2021physicalai,bousetouane2025physicalaiagents}. This perspective is closely related to embodied AI and robotics, but it is broader than robot control alone. It includes any intelligent system whose decisions are expressed through physical action and whose learning is shaped by contact with real environments, sensor noise, actuation limits, safety envelopes, and material consequences.

The connection between Physical AI and IoAT is structural. Physical AI asks how an intelligent entity can act in the real world; IoAT asks how many such entities can be networked, coordinated, monitored, and governed across heterogeneous cyber-physical infrastructures. A mobile robot, autonomous vehicle, or surgical assistant may be a concentrated Physical AI system with a clear body. A smart building or factory, by contrast, has a distributed body made of HVAC equipment, locks, cameras, lights, meters, controllers, mobile robots, and human interfaces. IoAT provides the architecture through which this distributed body can be perceived, reasoned about, and acted upon by multiple agents.

This distinction matters because physical intelligence at infrastructure scale cannot be reduced to a single embodied model. It requires capability discovery, access control, digital-twin simulation, local fallback, shared memory, human oversight, and cross-agent coordination. IoAT therefore complements Physical AI by providing the network layer, governance layer, and workflow-control layer needed to turn physically grounded intelligence into reliable operation across buildings, hospitals, transportation systems, energy systems, and industrial facilities.

\subsection{Convergence Toward IoAT}

The Internet of Agentic Things emerges from the convergence of IoT, CPS, Physical AI, agentic AI, and multi-agent systems. In an IoAT architecture, IoT devices are not merely endpoints that report data or execute fixed commands. They are embedded in an agentic network that can perceive environmental conditions, reason about objectives, coordinate specialized services, execute physical actions, and adapt through monitored outcomes. Agents may reside on edge devices, gateways, cloud platforms, enterprise servers, or embedded controllers while continuously interacting with sensors and actuators.

The defining characteristic of IoAT is autonomous closed-loop operation. Sensing, reasoning, planning, execution, and feedback occur continuously, with human oversight reserved for policy definition, exception handling, and high-consequence decisions \cite{zhu2026ioai,gartner2024cps,zhu2024cyberresiliencefoundations}. Rather than functioning as passive data collection networks, IoAT systems become distributed intelligent ecosystems that adapt to environmental changes and operational objectives in real time.

Several emerging frameworks illustrate this vision. Zhu's Internet of Agentic AI framework implicitly extends to IoT environments by enabling agents to operate across cloud, enterprise, edge, and cyber-physical domains \cite{zhu2026ioai}. Autonomous vehicles provide a representative example: LiDAR, radar, cameras, maps, prediction modules, planning modules, and control modules form an agentic perception-action loop that evaluates risk and generates physical control actions. Zhang et al. propose agent-driven orchestration mechanisms for next-generation communication networks, where high-level objectives such as mobility-aware vehicle-to-everything coordination or energy-efficient IoT management are decomposed into coordinated workflows spanning radio networks, edge resources, and control infrastructure \cite{zhang2026orchestration}. These examples show how abstract goals can be transformed into coordinated multi-agent actions across heterogeneous systems.

Industrial and infrastructure applications further demonstrate the practical relevance of IoAT. Predictive maintenance uses IoT sensors to monitor equipment conditions while AI agents analyze operational patterns, anticipate failures, and adapt maintenance schedules \cite{csg2025agentic}. Smart buildings, energy systems, and utility infrastructures can use agentic orchestration to optimize resource allocation, energy consumption, access control, and safety procedures beyond the capabilities of traditional rule-based automation. Cybersecurity is another important domain. Nageshwaran et al. envision self-defending IoT ecosystems in which LLM-enabled agents collaborate to monitor network behavior, interpret anomalies, explain threats, and synthesize response strategies in real time \cite{nageshwaran2026agentic}. Such systems could reconfigure network resources, isolate compromised devices, and deploy mitigation strategies while keeping human operators informed.

The emerging literature suggests a hierarchical but interactive deployment pattern. Device-level agents support local sensing, monitoring, and low-latency control actions; edge agents provide regional coordination, data fusion, local reasoning, and fallback behavior; cloud and enterprise agents provide strategic planning, optimization, large-scale learning, cross-system orchestration, and access to specialized knowledge \cite{zhang2026orchestration,mdpi2024agentreview,zhu2026ioai}. Coordination mechanisms from multi-agent systems, including negotiation, consensus formation, task allocation, coalition formation, and distributed planning, can operate over IoT communication infrastructures while respecting latency, privacy, safety, incentive, and trust constraints \cite{ibm2025agent,yangzhu2026ioaiworkflow,basar2026sociotechnical}. Conceptually, IoAT extends CPS by distributing intelligence and autonomy throughout the network, and it extends IoT by adding reasoning, planning, learning, and adaptive control as first-class architectural functions.

%% file: internet_agents.tex
The Internet of AI agents can be understood as a shift from monolithic intelligence to distributed, networked intelligence. In the monolithic view, a user interacts with a single assistant that is expected to contain all relevant knowledge, tools, memory, and reasoning ability. In the networked view, many autonomous or semi-autonomous agents coexist. Each agent may have specialized capabilities such as planning, coding, diagnosis, negotiation, sensing, simulation, security analysis, or compliance checking. Agents are connected through communication protocols and can dynamically interact to accomplish tasks that no single model or local system can handle efficiently. This view is consistent with recent work on language agents that interleave reasoning and action, learn tool use, and use foundation models as general-purpose substrates for autonomous agents, as well as with the Internet of Agentic AI view of large-scale communication, coordination, and collective intelligence among heterogeneous agents \cite{yao2023react,schick2023toolformer,xi2023rise,zhu2026ioai}.

\begin{figure}[t!]
    \centering
    \includegraphics[width=0.98\textwidth]{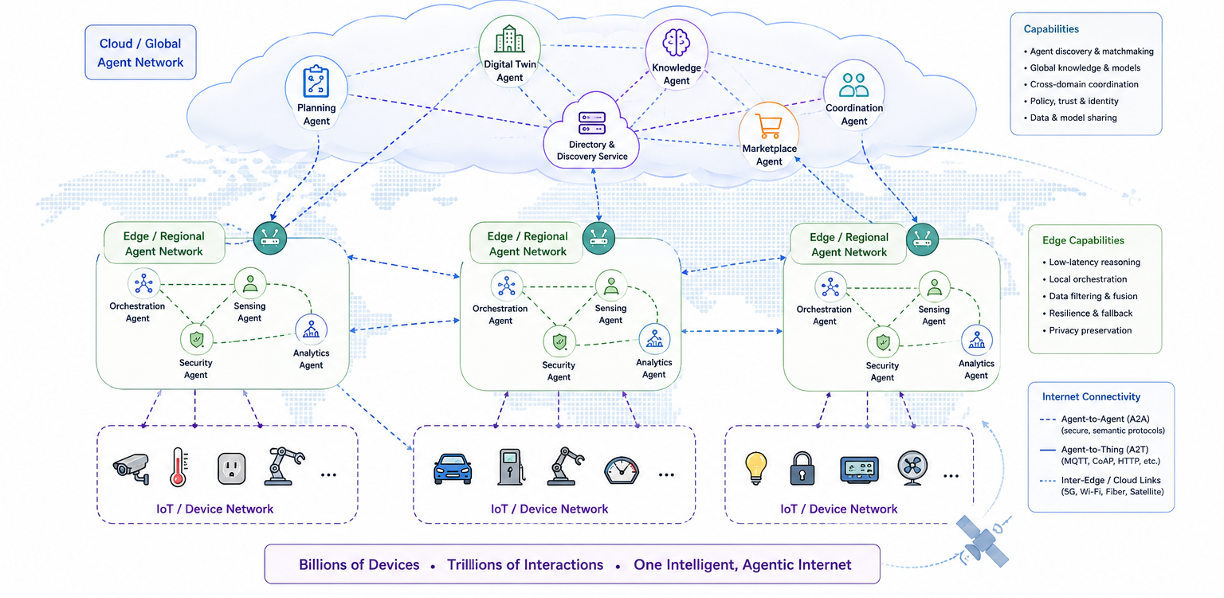}
    \caption{Network view of IoAT. Cloud and global agent networks provide planning, digital twins, knowledge, discovery, coordination, and marketplace services, while regional edge agent networks connect to local IoT device networks through secure agent-to-agent, agent-to-thing, and inter-edge links.}
    \label{fig:ioat-network}
\end{figure}

\begin{figure}[t!]
    \centering
    \includegraphics[width=0.98\textwidth]{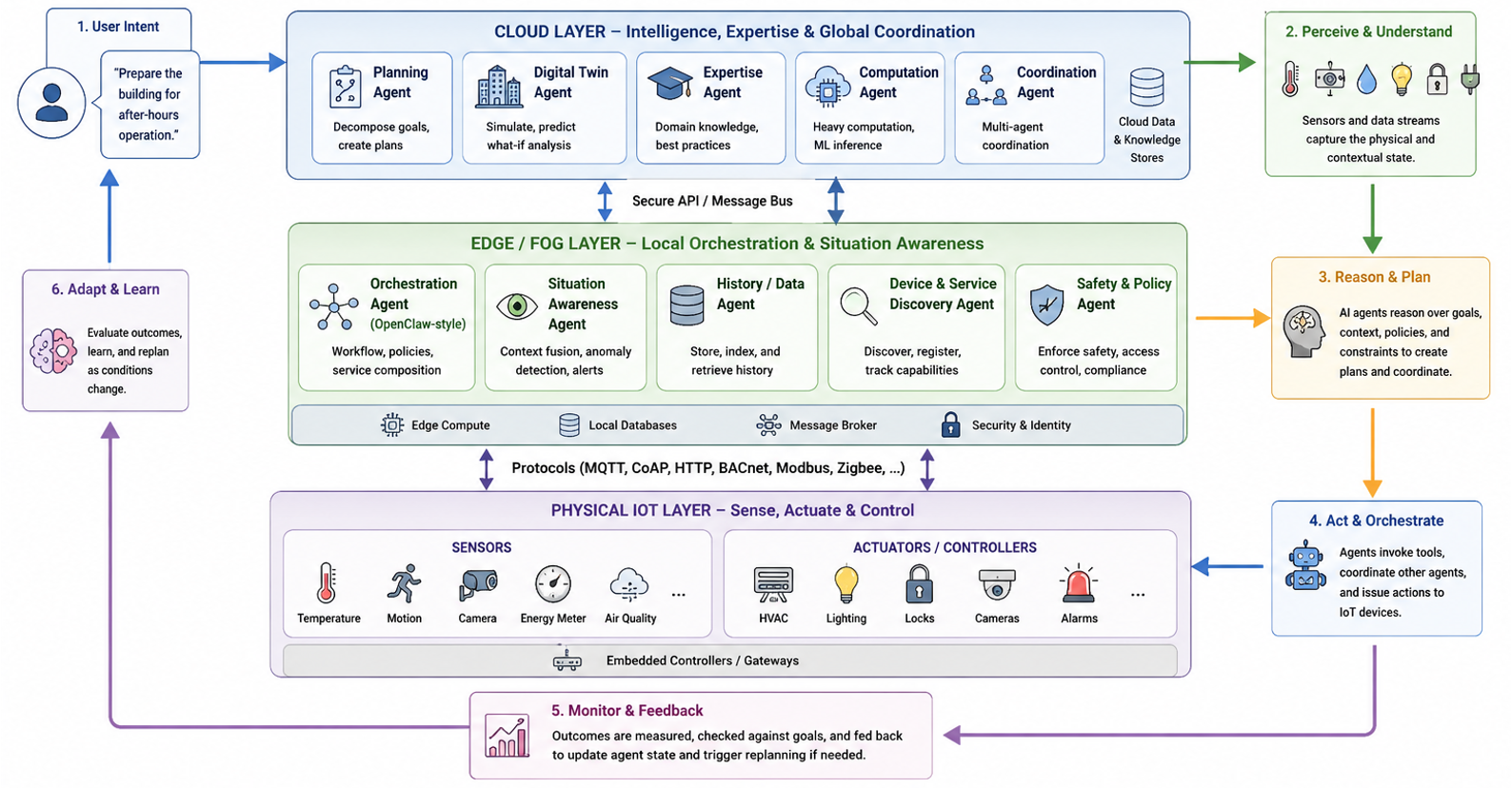}
    \caption{Conceptual overview of the Internet of Agentic Things. A user intent is translated into cloud-level planning, expertise, computation, and coordination; edge/fog agents provide local orchestration, situation awareness, history, service discovery, and safety-policy enforcement; and the physical IoT layer senses, actuates, monitors, and feeds outcomes back for adaptation and learning.}
    \label{fig:ioat-overview}
\end{figure}

This networked view is important because realistic agent capabilities are distributed. Some agents rely on proprietary data, specialized infrastructure, institutional context, local sensors, or domain-specific expertise. They may only be available at certain cloud services, organizational servers, edge gateways, or embedded devices. Consequently, agents must often be accessed and composed over a network rather than instantiated locally, and coalition formation becomes a practical coordination problem rather than only an implementation detail \cite{yangzhu2026ioaiworkflow}. Figure~\ref{fig:ioat-network} gives a network-level view of how cloud, edge, and device networks interact, while Figure~\ref{fig:ioat-overview} gives a process-level view of the closed-loop conceptual architecture.

The Internet of Agents provides several capabilities that become foundational for IoAT.

\medskip
\noindent\textbf{Capability routing and retrieval.} Agents can query the network for capabilities rather than assuming all expertise is local. A local building agent, for example, may invoke a weather-risk agent, a digital-twin simulation agent, or an energy-pricing agent only when those capabilities are needed.

\smallskip
\noindent\textbf{Service-based composition.} Complex tasks can be executed by chaining and composing agents across nodes. A task such as ``prepare the building for after-hours operation'' may require occupancy inference, security policy evaluation, energy optimization, access-control configuration, and notification generation. Each function may be provided by a different agent.

\smallskip
\noindent\textbf{Edge integration and locality.} IoAT requires agents at the edge. Local agents observe physical context, enforce real-time constraints, and preserve operational continuity when cloud services are delayed or unavailable. Edge agents can consume global services while providing local intelligence to the larger network.

\smallskip
\noindent\textbf{Shared memory and reuse.} Frequently requested outputs, validated plans, simulation results, and successful control policies can be stored and reused. This creates a distributed memory layer that reduces redundant computation and accelerates future response to similar scenarios.

\smallskip
\noindent\textbf{Network-level learning.} Over time, the system can improve by routing tasks to effective agents, reusing successful patterns, detecting failed compositions, and updating memory with execution outcomes. The network learns not only through model updates but also through better coordination, better caching, and better assignment of tasks to agents.

IoAT extends this Internet of Agents by coupling it to the physical world. Agents are not only exchanging text, code, or digital artifacts. They are producing control specifications, invoking IoT APIs, configuring devices, and monitoring physical consequences. This creates a hybrid network in which the agentic layer reasons about tasks and the IoT layer executes them under physical constraints.

This is where IoAT connects directly to Physical AI. Physical AI emphasizes intelligent action in real environments, often through robots, machines, vehicles, or other embodied systems. IoAT generalizes this idea from a single embodied agent to a distributed cyber-physical network. A smart building, for example, does not have one body in the way a robot does; its body is distributed across sensors, doors, cameras, HVAC equipment, lights, alarms, edge gateways, and human interfaces. The Internet of Agents supplies the coordination logic, while the IoT layer supplies the distributed physical embodiment. The result is a networked form of Physical AI in which physical action is planned, delegated, monitored, and adapted across many connected things.

Figure~\ref{fig:ioat-overview} summarizes this transition. Traditional IoT is centered on data collection and monitoring. IoAT adds perception, reasoning, action, and adaptation. It combines cloud agents for heavy reasoning and reusable computation, edge agents for tactical execution, and IoT devices for sensing and actuation. The result is a cyber-physical network in which every connected device can become part of an intelligent control and coordination loop.

%% file: architecture.tex
The proposed IoAT architecture has three coupled layers: a physical IoT layer, an edge/fog agentic layer, and a cloud agentic layer. This layering follows the broader movement from centralized cloud processing toward edge and fog architectures for latency-sensitive IoT applications \cite{bonomi2012fog,shi2016edge,satyanarayanan2017emergence}. It also clarifies IoAT's relationship to Physical AI: the physical layer provides the distributed body, the edge/fog layer provides local agency and situated control, and the cloud layer provides reusable reasoning, simulation, and coordination. The architecture separates strategic reasoning from tactical execution while preserving closed-loop feedback between decisions and physical outcomes.

\begin{figure}[t!]
    \centering
    \includegraphics[width=0.94\textwidth]{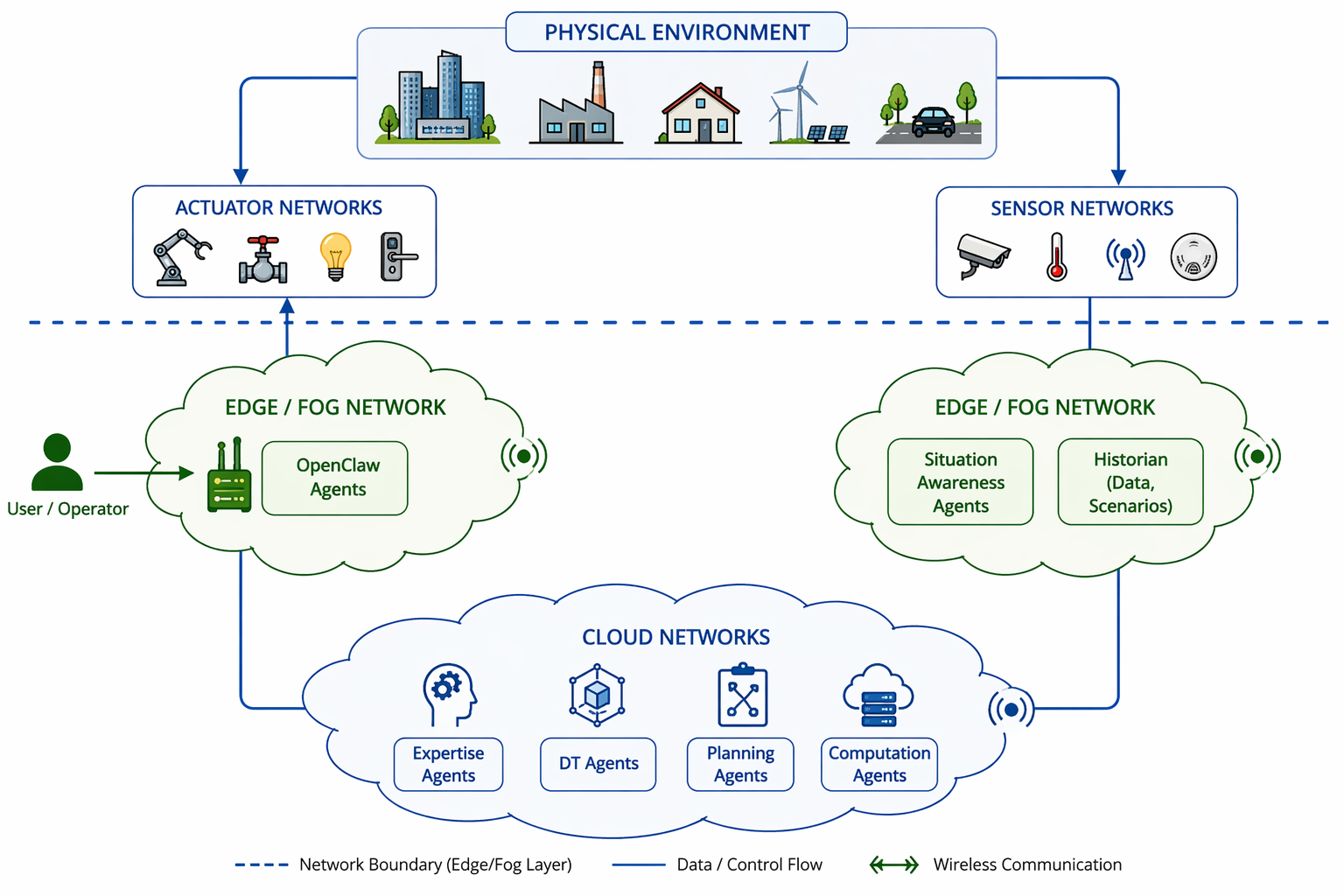}
    \caption{Hierarchical IoAT architecture. The physical environment is observed by sensor networks and influenced by actuator networks. Edge/fog agents provide local orchestration, situation awareness, and memory. Cloud agents provide planning, digital twins, domain expertise, and computation.}
    \label{fig:hierarchical}
\end{figure}

\subsection{Physical IoT Layer}

The physical layer contains the environment, sensor networks, actuator networks, and embedded controllers. Sensors observe state variables such as temperature, occupancy, access events, air quality, camera detections, equipment health, and energy consumption. Actuators influence the environment through devices such as HVAC controllers, valves, motors, lights, locks, alarms, and robotic mechanisms. Embedded controllers provide low-level feedback loops and safety interlocks. From a Physical AI perspective, this layer is the embodiment through which intelligence touches the world. In a robot, embodiment may be concentrated in a single machine; in IoAT, embodiment is distributed across many connected devices and physical subsystems.

This layer is governed by physical dynamics, communication constraints, safety limits, and device-level failure modes. It cannot be treated as a stateless API surface. A command issued by an agent changes the physical state, consumes energy, may affect occupants, and may interact with other control loops. For this reason, IoAT requires feedback-aware control rather than one-shot automation.

\subsection{Edge/Fog Agentic Layer}

The edge/fog layer hosts agents that require locality, low latency, and direct access to device networks. The central agent in the slide framework is the OpenClaw agent. OpenClaw acts as an agentic controller that receives user or application intent, maintains context, invokes other agents, coordinates IoT APIs, and monitors execution. It can combine LLM-driven reasoning with tool integration, memory, adaptive execution, and model predictive control concepts \cite{yao2023react,schick2023toolformer,mayne2000mpc,rawlings2017mpc}. In the Physical AI interpretation, OpenClaw is not itself the whole body of the system; it is a situated nervous system that coordinates perception and actuation across a distributed physical body.

OpenClaw is supported by two additional edge functions. Situation-awareness agents collect and interpret sensor data, converting raw telemetry into operational context such as ``building occupied,'' ``equipment fault,'' ``security anomaly,'' or ``temperature outside comfort band.'' Historian or data agents store logs, scenarios, prior plans, validated simulation outputs, and execution outcomes. These agents enable reuse: if a current situation matches a previous scenario, the system can retrieve a validated plan instead of recomputing from scratch.

The edge layer is also the right place for policy enforcement. It can restrict which tools agents may call, validate commands against device limits, require human approval for sensitive actions, and provide fallback control when cloud agents are unavailable.

\subsection{Cloud Agentic Layer}

The cloud layer provides heavy, global, and reusable intelligence. Planning agents decompose high-level goals into workflows, allocate subtasks, and reason over temporal, logical, and resource constraints. Digital-twin agents simulate physical outcomes under candidate strategies and provide a safer environment for evaluating plans before execution \cite{tao2019digitaltwin}. Expertise agents provide domain-specific knowledge, such as building operations, cybersecurity, emergency procedures, or regulatory constraints. Computation agents support optimization, large-scale search, scheduling, and model training. For Physical AI, this layer can provide world models, skill libraries, policy evaluation, and cross-site learning, while the edge layer decides which cloud capabilities are safe and timely enough to apply locally.

The cloud layer is not expected to make every real-time decision. Its strength is strategic reasoning, computation, and reuse across sites. The edge layer uses cloud outputs as plans, policies, models, or recommendations, and then adapts them to local conditions during execution.

\subsection{Cross-Layer Data and Control Flows}

Figure~\ref{fig:hierarchical} emphasizes that IoAT is not a simple top-down hierarchy. Information and control flow in both directions. Sensors provide telemetry and events to situation-awareness agents. Edge agents query cloud planners and digital twins. Cloud agents return workflows, policies, or control specifications. Edge agents execute commands through IoT APIs. Physical outcomes are observed, logged, evaluated, and folded back into memory.

This cross-layer structure creates five architectural principles.

The first principle is locality. Time-sensitive decisions should remain close to the devices and physical environment that they affect. In an IoAT architecture, edge agents should not function only as passive relays between sensors and the cloud. They should maintain enough contextual awareness and operational authority to execute safe local actions, reject commands that violate local constraints, and preserve continuity when wide-area connectivity is delayed, degraded, or unavailable. Locality is especially important for safety-critical actuation, privacy-sensitive sensing, and rapid anomaly response, because the cost of waiting for remote reasoning can be physical rather than merely computational.

The second principle is compositionality. Complex cyber-physical tasks should be decomposed into agent services with explicit interfaces, assumptions, and responsibilities. A high-level objective such as preparing a building for after-hours operation is not a single device command. It requires occupancy interpretation, access-control decisions, HVAC scheduling, lighting control, alarm configuration, notification, and verification. IoAT should therefore expose these functions as composable agentic services whose inputs, outputs, permissions, latency expectations, and failure modes are visible to the orchestrating agent. Compositionality makes the system scalable because new devices, agents, and domain services can be introduced without redesigning the entire architecture.

The third principle is feedback. Every plan must be monitored against physical outcomes and revised when conditions change. An agentic layer may generate a plausible workflow, but the physical layer determines whether the workflow is actually feasible, safe, and effective. Sensor streams, device acknowledgements, residual errors, alarms, and human interventions should be folded back into the workflow state so that the system can distinguish nominal execution from drift, fault, or adversarial manipulation. Feedback turns IoAT from open-loop automation into closed-loop cyber-physical intelligence.

The fourth principle is reuse. Validated plans, simulations, histories, and execution summaries should become reusable assets. Many IoAT environments encounter recurring operational patterns such as after-hours building preparation, seasonal HVAC operation, equipment-fault response, routine security checks, and emergency-mode transitions. When a plan has been evaluated in a digital twin and then validated through physical execution, the history agent can store the conditions under which it succeeded, the costs it incurred, and the exceptions it encountered. Future agents can retrieve, adapt, and revalidate prior solutions instead of recomputing from scratch, improving efficiency while also making behavior more predictable.

The fifth principle is governance. Agent actions must be constrained by access control, safety policies, auditability, and human oversight. Because IoAT agents can affect physical devices, their decisions must be treated as operational acts rather than only recommendations. Each tool call and device command should be checked against authorization, safety envelopes, policy constraints, and escalation rules. Governance also requires traceable records of which agent proposed an action, which evidence supported it, which policy allowed it, and what physical outcome followed. This accountability layer is essential for trustworthy deployment in buildings, hospitals, transportation systems, industrial facilities, and other high-consequence environments.

Together, these principles move IoT from passive connectivity toward active, adaptive, and accountable cyber-physical intelligence. They also clarify why IoAT cannot be reduced to adding a language-model interface on top of existing device APIs. The architecture must distribute reasoning across cloud and edge agents, maintain a live feedback channel to the physical layer, preserve reusable operational memory, and enforce governance wherever agentic reasoning becomes physical action. This is the architectural condition for trustworthy Physical AI at infrastructure scale.

%% file: usecase.tex
A concrete use case helps illustrate how IoAT differs from conventional IoT automation. Consider a commercial smart building that integrates smart locks, HVAC controllers, cameras, lighting, alarms, energy meters, and environmental sensors. A user issues the command:
\[
\text{``Prepare the building for after-hours operation.''}
\]
In a conventional IoT system, this request might require a human operator to inspect dashboards and manually trigger several automations. In IoAT, the request becomes an agentic workflow.

The same scenario can also be read as a Physical AI example. The building is not a humanoid robot, but it has a distributed physical body: sensors perceive occupancy and environmental state, controllers and actuators change HVAC, lighting, locks, alarms, and access conditions, and edge agents coordinate these changes under physical and policy constraints. IoAT provides the networked agency that allows this distributed body to act coherently in response to high-level intent.

\begin{figure}[t!]
    \centering
    \includegraphics[width=0.94\textwidth]{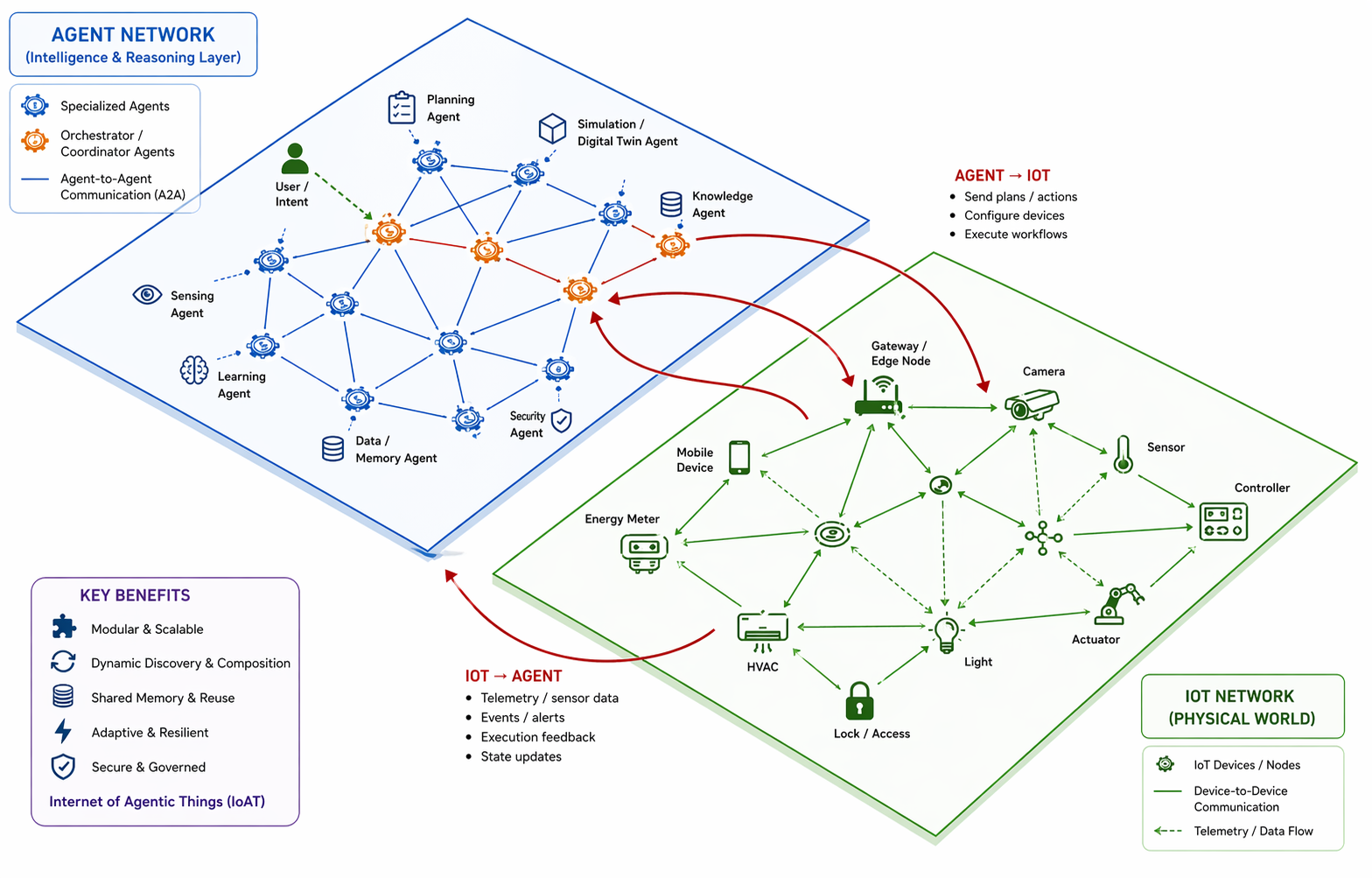}
    \caption{Two-layer IoAT network. The agent network provides intelligence, reasoning, planning, simulation, knowledge, and coordination. The IoT network provides cameras, sensors, HVAC, locks, lights, controllers, actuators, and edge gateways. Red arrows represent cross-layer interaction between agentic decisions and physical execution.}
    \label{fig:two-layer}
\end{figure}

\subsection{Intent Interpretation and Context Gathering}

OpenClaw first interprets the user intent. The phrase ``after-hours operation'' is not a single device command. It implies a policy state for the building: occupants should be protected, energy should be conserved, access should be controlled, and security monitoring should be enabled. OpenClaw queries situation-awareness agents to determine the current context. Relevant observations include occupancy, scheduled events, open doors, security zones, outdoor temperature, HVAC state, camera status, maintenance exceptions, and energy-pricing conditions.

If the historian contains a validated plan for a similar after-hours scenario, OpenClaw can retrieve it and adapt parameters. If the current context is novel or uncertain, OpenClaw invokes planning agents and digital-twin agents to generate and evaluate candidate plans.

\subsection{Workflow Decomposition}

The high-level request is decomposed into coordinated subtasks. The workflow first verifies occupancy and detects whether any people remain in the building. It then adjusts HVAC to an energy-saving mode while preserving safety and equipment constraints, dims or turns off lighting in unoccupied zones, locks doors while configuring access-control exceptions, activates cameras, alarms, and monitoring policies, and notifies operators if anomalies or unresolved exceptions remain.

Each subtask may be assigned to a specialized agent or tool. The planning agent determines precedence constraints. For example, access doors should not be fully locked until occupancy checks are complete. HVAC setbacks should account for expected night temperature, equipment limits, and next-day schedules. Security alarms should be activated only after authorized maintenance or cleaning exceptions are checked.

\subsection{Physical Execution and Replanning}

Once a plan is selected, OpenClaw coordinates execution through IoT APIs. The device network may include gateways, controllers, mobile interfaces, cameras, energy meters, HVAC devices, locks, lights, sensors, and actuators, as illustrated in Figure~\ref{fig:two-layer}. During execution, the system monitors whether the physical state follows the expected trajectory.

Unexpected events trigger replanning. If a camera or occupancy sensor detects a person in the building, the workflow may pause door locking, maintain lighting in occupied zones, and notify security staff. If an HVAC fault is detected, the agent may change the energy plan, open a maintenance ticket, and adjust comfort constraints. If a security breach is detected, the system may activate alarms, lock specific zones, preserve video evidence, and escalate to human operators.

\subsection{Value of the IoAT Approach}

This use case highlights four advantages of IoAT. First, natural language is translated into structured, multi-step physical control. Second, specialized agents contribute distinct capabilities without requiring a single monolithic controller. Third, histories and digital twins support reuse and safer evaluation of plans. Fourth, physical feedback closes the loop, allowing the system to adapt when reality deviates from the nominal plan.

The same pattern extends beyond smart buildings and connects directly to Physical AI. In healthcare facilities, IoAT agents could coordinate clinical devices, facility systems, robotic assistants, and emergency procedures. In manufacturing, they could coordinate production lines, robots, sensors, and maintenance systems. In transportation, they could coordinate vehicles, traffic signals, infrastructure sensors, and incident response. The key idea is the same: high-level intent becomes distributed agentic planning, and agentic planning becomes closed-loop physical execution.

%% file: workflow_control.tex
IoAT requires a model that connects agentic workflow design with physical execution. The smart-building example in Section~\ref{sec:usecase} illustrates why such a model is needed: a request such as ``prepare the building for after-hours operation'' is simultaneously a planning problem, a multi-agent coordination problem, a Physical AI problem, and a physical control problem. The central difficulty is the coexistence of two time scales. At the strategic time scale, an agentic network interprets user intent, selects a workflow, allocates agents, and configures constraints. At the tactical time scale, edge controllers and IoT devices execute the resulting specifications through sensing, actuation, feedback, and local control. In Physical AI terms, the strategic layer forms an intention for a distributed body, while the tactical layer determines whether that intention can be safely realized through physical motion, switching, flow, access, and energy. This section formalizes the coupling using a hylomorphic dynamic-programming viewpoint inspired by fold-unfold recursion schemes and hylomorphic dynamic programming \cite{meijer1991functional,hutton1998fold,hutton1999tutorial,spivak2014category,yangzhu2026hdp}: strategic configurations are \emph{unfolded} into tactical consequences, and tactical outcomes are then \emph{folded} back into strategic value, memory, and replanning. The same two-level perspective is compatible with secure and resilient CPS control, where planning must account for adversarial actions, switching among controllers, recovery stages, and strategic interaction between attackers and defenders \cite{miao2017movinghorizon,zhu2024cyberresiliencefoundations}.

\subsection{Strategic and Tactical Time Scales}

Let $n=0,1,\ldots,N$ index strategic epochs and let $k=0,1,\ldots,K_n$ index tactical steps within epoch $n$. The strategic state
\[
Z_n\in \cZ_n
\]
summarizes the information available to the agentic layer. In IoAT, $Z_n$ may include user intent, task-graph state, global situational estimates, historian memory, active policies, available agents, device-health summaries, and risk posture. Given $Z_n$, the agentic layer chooses a strategic configuration
\[
\theta_n\in \Theta_n(Z_n).
\]
The configuration $\theta_n$ may specify a workflow decomposition, an assignment of subtasks to agents, an operating mode, a control objective, a constraint set, an access policy, a digital-twin query, or a replanning rule.

The tactical layer executes the configuration. Let
\[
X_{n,k}\in \cX,\qquad u_{n,k}\in \cU_n(X_{n,k}),\qquad w_{n,k}\in \cW
\]
denote the physical IoT state, control action, and disturbance at tactical step $k$ of epoch $n$. The tactical episode is initialized by the strategic configuration,
\[
X_{n,0}=X_{\mathrm{init}}(Z_n,\theta_n),
\]
and evolves according to
\[
X_{n,k+1}=f_n(X_{n,k},u_{n,k};Z_n,\theta_n,w_{n,k}).
\]
This dependence on $(Z_n,\theta_n)$ captures the way the agentic layer shapes local execution. For example, an after-hours building workflow may initialize the tactical episode with the current occupancy estimate and then impose HVAC, lighting, lock, and camera constraints.

The tactical execution produces a trajectory
\[
\tau_n=\{(X_{n,k},u_{n,k})\}_{k=0}^{K_n}
\]
and an outcome summary
\[
\sigma_n=\Psi_n(\tau_n)\in \Sigma.
\]
The summary $\sigma_n$ may contain physical outcomes, constraint violations, execution cost, alerts, anomalies, residual errors, or a compact success/failure report. The strategic state then evolves according to the controlled kernel
\[
Z_{n+1}\sim G_n(\cdot\mid Z_n,\theta_n,\sigma_n).
\]
Thus, the slow state update depends not only on the strategic action but also on what actually happened during physical execution.

\subsection{Tactical Layer as an Inner Dynamic Program}

For a fixed strategic configuration $(Z_n,\theta_n)$, the tactical layer solves an embedded finite-horizon control problem. Let the tactical step cost and terminal cost be
\[
\ell_n(X_{n,k},u_{n,k};Z_n,\theta_n),\qquad
\varphi_n(X_{n,K_n};Z_n,\theta_n).
\]
The tactical value is
\[
J_{\mathrm{in},n}(Z_n,\theta_n)
=
\min_{\pi_{\mathrm{in},n}}
\mathbb{E}
\left[
\sum_{k=0}^{K_n-1}
\ell_n(X_{n,k},u_{n,k};Z_n,\theta_n)
+\varphi_n(X_{n,K_n};Z_n,\theta_n)
\right],
\]
where $u_{n,k}=\pi_{\mathrm{in},n,k}(X_{n,k};Z_n,\theta_n)$ and the expectation is taken over disturbances and any stochasticity in the physical system.

The corresponding Bellman recursion is
\[
V_{\mathrm{in},n,K_n}(x;Z_n,\theta_n)
=
\varphi_n(x;Z_n,\theta_n),
\]
and, for $k=K_n-1,\ldots,0$,
\[
\begin{aligned}
V_{\mathrm{in},n,k}(x;Z_n,\theta_n)
=
\min_{u\in \cU_n(x)}
\Big[
&\ell_n(x,u;Z_n,\theta_n) \\
&+
\mathbb{E}_{w}
V_{\mathrm{in},n,k+1}
\big(f_n(x,u;Z_n,\theta_n,w);Z_n,\theta_n\big)
\Big].
\end{aligned}
\]
The optimized tactical value is
\[
J_{\mathrm{in},n}(Z_n,\theta_n)
=
V_{\mathrm{in},n,0}(X_{\mathrm{init}}(Z_n,\theta_n);Z_n,\theta_n).
\]
The optimal tactical policy induces a distribution over outcome summaries,
\[
P^{\Psi}_n(\sigma\mid Z_n,\theta_n)
=
\mathbb{P}\{\Psi_n(\tau_n)=\sigma\mid Z_n,\theta_n,\pi_{\mathrm{in},n}^{*}\}.
\]
This distribution is the statistical interface from the physical IoT layer back to the agentic layer.

\subsection{Anamorphism: Unfolding Agentic Configurations}

The inner dynamic program defines the anamorphic, or unfold, operator. For each epoch $n$,
\[
\mathsf{A}_n(Z_n,\theta_n)
=
\Big(
J_{\mathrm{in},n}(Z_n,\theta_n),
P^{\Psi}_n(\cdot\mid Z_n,\theta_n)
\Big).
\]
The operator $\mathsf{A}_n$ expands a compact strategic configuration into its tactical consequences. In IoAT, this means that a high-level agentic decision such as ``secure the building while minimizing energy use'' is unfolded into device trajectories, actuation costs, comfort deviations, security events, and a distribution over execution summaries.

\begin{figure}[t!]
    \centering
    \includegraphics[width=0.94\textwidth]{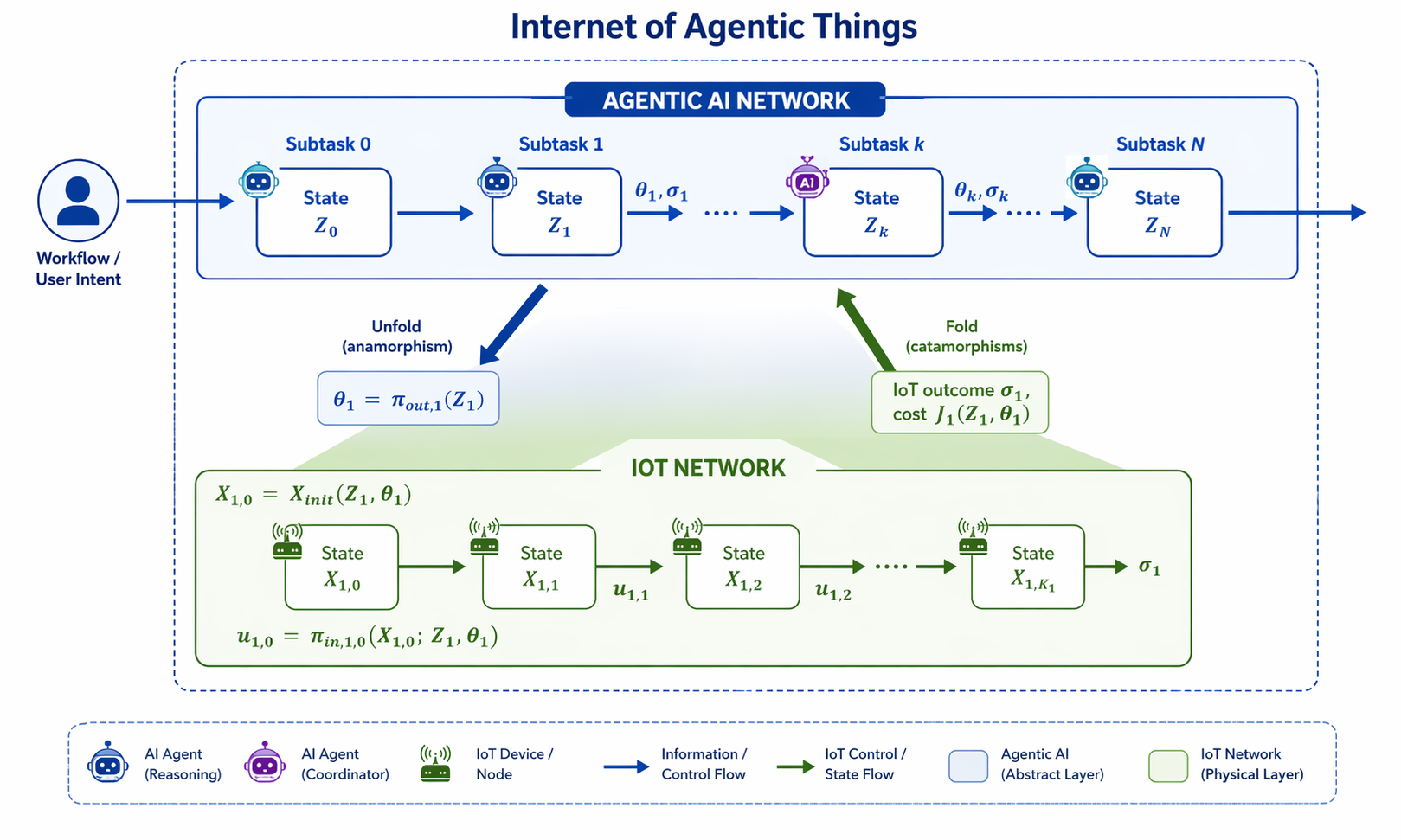}
    \caption{Hylomorphic workflow-control coupling in IoAT. User intent and strategic agent state are unfolded into tactical IoT episodes. Physical outcomes, costs, and summaries are folded back into strategic value, memory, and replanning.}
    \label{fig:workflow-control}
\end{figure}

Figure~\ref{fig:workflow-control} illustrates this fold-unfold interpretation. The upper layer tracks the abstract agentic workflow state $Z_n$, while the lower layer executes the physical IoT process conditioned on $(Z_n,\theta_n)$. The anamorphism maps $(Z_n,\theta_n)$ into an optimized tactical episode and a summary distribution.

\subsection{Catamorphism: Folding Tactical Outcomes}

The catamorphic, or fold, operator aggregates tactical consequences into the strategic value recursion. There are two natural information regimes.

In the outcome-observable regime, the agentic layer observes the tactical summary $\sigma_n$. The strategic value recursion is
\[
V^{o}_{\mathrm{out},N}(Z_N)=\Phi_N(Z_N),
\]
and, for $n=N-1,\ldots,0$,
\[
\begin{aligned}
V^{o}_{\mathrm{out},n}(Z_n)
=
\min_{\theta\in \Theta_n(Z_n)}
\Big[
&J_{\mathrm{in},n}(Z_n,\theta) \\
&+
\mathbb{E}_{\sigma\sim P^{\Psi}_n(\cdot\mid Z_n,\theta)}
\mathbb{E}_{Z'\sim G_n(\cdot\mid Z_n,\theta,\sigma)}
V^{o}_{\mathrm{out},n+1}(Z')
\Big].
\end{aligned}
\]
This case is appropriate when the edge layer reports a trusted execution summary to the agentic layer, such as a verified occupancy result, a completed lock-down status, or an anomaly report.

In the outcome-unobservable regime, the agentic layer does not directly observe $\sigma_n$ and must reason through the effective transition kernel
\[
\bar{G}_n(Z'\mid Z_n,\theta)
=
\int_{\Sigma}
G_n(Z'\mid Z_n,\theta,\sigma)
P^{\Psi}_n(d\sigma\mid Z_n,\theta).
\]
The strategic value recursion becomes
\[
V^{u}_{\mathrm{out},N}(Z_N)=\Phi_N(Z_N),
\]
and
\[
V^{u}_{\mathrm{out},n}(Z_n)
=
\min_{\theta\in \Theta_n(Z_n)}
\left[
J_{\mathrm{in},n}(Z_n,\theta)
+
\mathbb{E}_{Z'\sim \bar{G}_n(\cdot\mid Z_n,\theta)}
V^{u}_{\mathrm{out},n+1}(Z')
\right].
\]
This case is relevant when communication is delayed, summaries are unavailable, privacy constraints restrict reporting, or the agentic layer only receives coarse state updates.

The fold operator can be written abstractly as
\[
\mathsf{C}_n[V_{\mathrm{out},n+1}](Z_n)
=
V_{\mathrm{out},n}(Z_n),
\]
where the precise form of $\mathsf{C}_n$ depends on whether the outcome summary is observable or folded into the effective kernel.

\subsection{Hylomorphic IoAT Dynamic Programming}

The IoAT workflow-control problem is hylomorphic when the strategic value functions are generated by recursive composition of the unfold and fold operators:
\[
V_{\mathrm{out},n}
=
\mathsf{C}_n[V_{\mathrm{out},n+1}]\circ \mathsf{A}_n,
\qquad
n=N-1,\ldots,0,
\]
with terminal value $V_{\mathrm{out},N}=\Phi_N$. More explicitly, the agentic layer first evaluates each candidate configuration by unfolding it through $\mathsf{A}_n$, and then folds the resulting tactical value and summary law into the outer Bellman recursion through $\mathsf{C}_n$.

\begin{definition}[Hylomorphic IoAT Policy]
A policy pair $(\pi_{\mathrm{out}},\pi_{\mathrm{in}}^{*})$ is hylomorphic if, at every strategic epoch $n$, the tactical policy $\pi_{\mathrm{in},n}^{*}(\cdot;Z_n,\theta_n)$ solves the inner Bellman recursion for every feasible $(Z_n,\theta_n)$, and the strategic policy $\pi_{\mathrm{out},n}$ selects a minimizer of the corresponding outer fold recursion under the chosen information regime.
\end{definition}

\begin{proposition}[Dynamic Consistency of Hylomorphic IoAT Control]
Fix either the outcome-observable or outcome-unobservable information regime. If the tactical policies solve the inner Bellman recursions and the strategic policy solves the corresponding outer recursion, then the induced IoAT policy is dynamically consistent across strategic epochs: after every observed strategic state $Z_n$, the remaining policy is optimal for the continuation problem generated by the same unfold-fold operators.
\end{proposition}

The proposition follows directly from Bellman's principle applied first to each tactical episode and then to the strategic recursion. The important point for IoAT is modularity. The edge and physical layers can solve tactical episodes in parallel for candidate configurations, while the agentic layer uses only the folded outputs $J_{\mathrm{in},n}$ and $P^{\Psi}_n$ or $\bar{G}_n$ to select strategic actions. This avoids forcing the cloud-level planner to carry full device trajectories in its state, while still preserving the influence of physical execution on future agentic decisions.

\subsection{Application to Smart-Building After-Hours Operation}

The after-hours building scenario makes the abstract variables concrete. The strategic state $Z_n$ is the information state maintained by OpenClaw and its supporting agents at the beginning of a decision epoch. It contains the user intent, the current task status, the historian's record of similar after-hours episodes, the occupancy belief over zones, scheduled events, outdoor conditions, energy prices, device-health summaries, access-control policies, maintenance exceptions, and the current security posture. From the perspective of Physical AI, $Z_n$ is the cognitive state of a distributed embodied system: it summarizes what the building believes about its own physical condition, what it is trying to achieve, and which forms of action remain permissible. In compact form, one may write
\[
Z_n=(g_n,b_n,h_n,p_n,a_n,r_n),
\]
where $g_n$ denotes the active goal, $b_n$ the fused physical belief, $h_n$ the relevant history, $p_n$ the policy context, $a_n$ the available agent and device capabilities, and $r_n$ the current risk state. This state is not a raw sensor dump. It is the agentic summary that allows the system to reason about what it means to be ready for after-hours operation.

A strategic configuration $\theta_n$ is a candidate after-hours workflow. It may specify which agents are invoked, which checks must precede actuation, which digital-twin simulations are required, which device commands are permitted, which human approvals are needed, and which replanning thresholds are active. For the building example, OpenClaw may compare a conservative security-first configuration, an energy-saving configuration with deeper HVAC setbacks, and a maintenance-aware configuration that delays alarm activation in zones where authorized work is scheduled. Each configuration is therefore not merely a label such as ``secure mode.'' It is a structured workflow with precedence constraints, access policies, local control objectives, and fallback rules. When required capabilities are distributed across separate organizations, cloud services, or edge domains, this configuration can also be interpreted as a coalition-selection and workflow-allocation decision \cite{yangzhu2026ioaiworkflow}.

Once $\theta_n$ is selected for evaluation, the anamorphic step unfolds it into a tactical episode. The physical state $X_{n,k}$ can include zone occupancy, temperatures, humidity, air quality, lock states, lighting levels, alarm status, camera availability, power draw, and fault indicators. The tactical action $u_{n,k}$ can include HVAC setpoints, lighting commands, lock commands, camera and alarm activation, notification actions, and requests for local verification. These variables are the formal counterpart of the distributed body described in the use case: they specify what the building can sense, what it can change, and which material consequences follow from a proposed plan. The disturbance $w_{n,k}$ captures the uncertainties that make the problem genuinely cyber-physical: a late occupant may remain in a room, an exterior door may fail to lock, outdoor temperature may change faster than expected, a camera may become unavailable, or a network link may delay acknowledgement.

The inner cost $\ell_n$ expresses what the tactical layer is trying to accomplish under the strategic configuration. In the after-hours case, it can penalize energy consumption, comfort or safety violations, unsecured doors, lighting in empty zones, unauthorized access risk, delayed execution, excessive device cycling, and policy violations. The terminal cost $\varphi_n$ penalizes ending the episode in an unacceptable state, such as unresolved occupancy, an unlocked perimeter door, an unarmed security zone, or an HVAC state that risks equipment damage. Thus, $J_{\mathrm{in},n}(Z_n,\theta_n)$ is the predicted physical cost of carrying out a candidate agentic workflow, not merely the computational cost of planning it.

The outcome summary $\sigma_n=\Psi_n(\tau_n)$ is the bridge back to the agentic layer. For the after-hours workflow, $\sigma_n$ may report that the building is clear and secured, that a person remains in a specific zone, that a device fault prevented full execution, that a maintenance exception requires human confirmation, or that the energy-saving plan succeeded with a measured cost and residual risk. In the outcome-observable regime, OpenClaw receives this summary from the edge layer and updates the strategic state through $G_n$. In the outcome-unobservable regime, perhaps because a communication link is delayed or privacy policy restricts detailed telemetry, OpenClaw reasons through the effective transition kernel $\bar{G}_n$ and chooses a more conservative continuation policy.

This application also clarifies the role of the digital twin and historian. Before executing a workflow, the digital-twin agent can estimate the distribution $P^{\Psi}_n(\cdot\mid Z_n,\theta_n)$ for each candidate configuration by simulating occupancy uncertainty, thermal dynamics, device failures, and security-policy interactions. The historian can bias this evaluation with prior after-hours episodes, identifying plans that previously succeeded under similar occupancy and weather conditions. The catamorphic step then folds these predicted tactical consequences into the outer strategic recursion. A plan that saves energy but has a high probability of leaving a zone insecure may be rejected in favor of a slightly more expensive plan with lower residual risk. Conversely, if the building has reliable occupancy clearance and mild weather, the folded value may favor an energy-saving configuration.

During execution, the same mapping explains replanning. If the realized summary $\hat{\sigma}_n$ reports that an occupant remains in a conference room, OpenClaw updates $Z_n$, pauses lock-down for the relevant zone, keeps lighting and ventilation active locally, and selects a new configuration that coordinates notification, access control, and security monitoring. If $\hat{\sigma}_n$ reports an HVAC fault, the new strategic state includes equipment risk and maintenance status, and the next configuration may trade energy savings for equipment protection and operator escalation. This is the workflow-control form of Physical AI: cognition is not complete until it is tested against the behavior of the physical system, and physical action is not complete until it is folded back into future cognition. In this sense, the smart-building example is not an informal illustration appended to the mathematics. It is an instance of the hylomorphic IoAT recursion: intent is unfolded into tactical device behavior, and physical outcomes are folded back into memory, risk, policy, and the next agentic decision.

\subsection{Illustrative Execution Scenarios}

The formal model becomes clearer when the variables are read as live execution states rather than as abstract symbols. In each scenario below, the same cycle is used. OpenClaw begins with a strategic state $Z_n$, evaluates a candidate configuration $\theta_n$ by unfolding it through the tactical IoT layer, receives either a predicted or realized summary $\sigma_n$, and folds that summary back into the next strategic state. The examples differ in which part of the framework dominates: perception and verification, digital-twin prediction, governance, or edge fallback.

\subsubsection{Occupied-Zone Exception}

Consider an after-hours workflow in which the current strategic goal $g_n$ is to secure the building, while the fused belief $b_n$ assigns high but not perfect probability to the building being empty. A conservative configuration $\theta_n^{\mathrm{sec}}$ instructs OpenClaw to verify occupancy before locking perimeter doors, arming alarms, and applying HVAC setbacks. The unfold operator maps this configuration into a tactical episode in which $X_{n,k}$ includes zone-level occupancy estimates, door states, lighting levels, alarm states, and HVAC modes, while $u_{n,k}$ includes commands to query sensors, keep lights on in uncertain zones, delay selected locks, and request camera confirmation.

During execution, suppose a motion sensor and camera agent report that a person remains in conference room $i$. The tactical summary can be written schematically as
\[
\hat{\sigma}_n=
(\mathrm{occupied\_zone}=i,\ \mathrm{lockdown}=\mathrm{incomplete},\ \mathrm{residual\_risk}=\mathrm{medium}).
\]
The inner cost $\ell_n$ assigns a high penalty to locking an occupied zone, turning off lights around the occupant, or applying an HVAC setback that violates comfort and safety constraints. The catamorphic step folds the realized summary into the next strategic state,
\[
Z_{n+1}\sim G_n(\cdot\mid Z_n,\theta_n^{\mathrm{sec}},\hat{\sigma}_n),
\]
so that the new state records an occupied-zone exception, a partially completed security workflow, and a higher residual risk. OpenClaw then selects a revised configuration that keeps lighting and ventilation active in room $i$, pauses alarm activation for the affected zone, notifies security or facility staff, and continues securing the rest of the building. This example illustrates why the model separates the high-level intention from the tactical episode: the goal remains after-hours readiness, but the physical evidence changes the feasible way to achieve it.

\subsubsection{HVAC Fault Under Energy-Saving Intent}

A second scenario begins with the user intent ``reduce overnight energy consumption while keeping the building ready for tomorrow morning.'' The strategic state contains weather forecasts, next-day schedules, utility prices, equipment health, and historian records of similar nights. OpenClaw may compare two candidate configurations. The first applies an aggressive setback, while the second applies a mild setback with earlier preconditioning. The digital-twin agent estimates the summary distributions
\[
P^{\Psi}_n(\cdot\mid Z_n,\theta_n^{\mathrm{deep}})
\quad \text{and} \quad
P^{\Psi}_n(\cdot\mid Z_n,\theta_n^{\mathrm{mild}}),
\]
where the summaries include energy cost, morning recovery time, comfort violations, and equipment stress.

If the historian indicates that a similar outdoor temperature previously caused slow morning recovery, the unfolded value $J_{\mathrm{in},n}(Z_n,\theta_n^{\mathrm{deep}})$ may be lower in energy cost but higher in terminal risk. If a supply-air sensor later reports an HVAC fault during execution, the realized summary contains both a physical fault and an incomplete energy plan. Folding this summary into $Z_{n+1}$ changes the active risk state $r_n$ and the available capability set $a_n$: the next configuration may abandon aggressive energy savings, protect equipment, open a maintenance ticket, and request operator confirmation before continuing. This scenario shows how the digital twin and historian enter the model as predictive components of the unfold step, while device faults enter the fold step as physical evidence that reshapes future planning.

\subsubsection{Security Alert and Governed Actuation}

A third scenario illustrates governance. Suppose the after-hours workflow detects a forced-door event near a service entrance. The strategic state now includes a security anomaly, an uncertain occupancy belief, and policies that constrain which doors may be locked, unlocked, or remotely overridden. A security-first configuration may instruct the edge layer to preserve camera evidence, lock nearby interior zones, activate alarms, notify operators, and prevent nonessential agents from issuing access-control commands. The tactical actions are not only physical commands; they are governed commands whose feasibility depends on access rights, reversibility, emergency-egress rules, and human-approval requirements.

The outcome summary may report that the perimeter door is open, two cameras are available, one lock failed to acknowledge, and no occupant is detected in the adjacent hallway. The fold step converts this summary into a new risk posture and a new policy context. In particular, a failed acknowledgement should not be treated as a successful lock command, and the absence of detected occupancy should not be treated as proof that the area is empty. The next strategic configuration may therefore combine containment with human escalation rather than full autonomous lock-down. This example shows how the model accommodates safety and trust: $\theta_n$ is not merely a plan for task completion, but a governed configuration whose admissible actions depend on policy, evidence quality, and physical consequence.

\subsubsection{Cloud Delay and Edge Fallback}

A fourth scenario illustrates the outcome-unobservable regime. Suppose OpenClaw requests a cloud planning agent and digital-twin evaluation, but wide-area connectivity becomes delayed. The edge layer cannot wait indefinitely because after-hours transitions involve locks, lights, alarms, and HVAC setpoints. In this case, OpenClaw uses a local fallback configuration $\theta_n^{\mathrm{edge}}$ that restricts actions to safe, reversible, and locally verifiable commands. It may keep occupied or uncertain zones in a conservative state, avoid deep HVAC setbacks, activate only alarms whose preconditions are locally verified, and postpone actions that require cloud expertise or human approval.

Because detailed summaries from some services are unavailable, the strategic layer reasons through the effective kernel $\bar{G}_n$ rather than through an observed $\sigma_n$. This does not mean the system acts blindly. It means that local summaries, prior histories, and conservative transition estimates replace the missing global summary until communication recovers. When cloud connectivity returns, the delayed digital-twin result and device acknowledgements are folded into the strategic memory, allowing OpenClaw to reconcile what was planned, what was executed locally, and what remains unresolved. This example illustrates the architectural principle of locality: the edge layer preserves safe operation when the cloud layer is useful but temporarily unavailable.

\subsection{Punctuated Equilibrium and Replanning}

The hylomorphic formulation also clarifies the role of punctuated equilibrium in agentic control. During nominal operation, the strategic layer executes the selected configuration and receives folded summaries from the tactical layer. Replanning is triggered when the realized observations deviate from the predicted tactical outcome distribution:
\[
d(\hat{\sigma}_n,P^{\Psi}_n(\cdot\mid Z_n,\theta_n))>\eta,
\]
or when physical observations leave an acceptable tube around the predicted trajectory. Here $d$ is a discrepancy measure, $\hat{\sigma}_n$ is the realized summary, and $\eta$ is an anomaly threshold.

A punctuation event initiates a new unfold-fold cycle. The agentic layer updates $Z_n$ using the new evidence, invokes a digital twin or planning agent if needed, unfolds candidate configurations through updated tactical models, and folds the resulting summaries into a revised strategic decision. Thus, replanning is not an exceptional add-on to IoAT; it is the operational manifestation of the hylomorphic recursion under uncertainty, disturbance, and incomplete information.

%% file: resilience.tex
IoAT systems inherit risks from agentic AI, cyber-physical IoT, and Physical AI. They must therefore be designed as reliable control systems, secure distributed systems, and accountable decision systems. The Physical AI connection raises the stakes because reasoning failures, corrupted context, or unsafe delegation can be expressed through locks, valves, vehicles, robots, alarms, lighting, clinical devices, or other physical interfaces. This perspective aligns with broader work on foundation-model risk, AI risk management, zero-trust architecture, systematic threat modeling, resilient control, and agentic cyber resilience \cite{bommasani2021opportunities,nist2023airmf,nist2020zero,shostack2014threat,miao2017movinghorizon,zhu2024cyberresiliencefoundations,lidzhu2025agenticcyberresilience}.

\subsection{Risk Pathways}

Several risk pathways are particularly important.

\medskip
\noindent\textbf{Prompt injection and tool manipulation.} Agents that consume text, logs, web content, tickets, or device metadata may encounter adversarial instructions. A malicious prompt embedded in a device name, maintenance note, or camera annotation could attempt to redirect the agent's behavior.

\smallskip
\noindent\textbf{Agent compromise and unsafe delegation.} If a specialized agent or tool is compromised, the orchestrator may receive incorrect recommendations or malicious actions. The risk increases when agents can call other agents recursively.

\smallskip
\noindent\textbf{Stale memory and invalid reuse.} Reusing historical plans is valuable, but it can be unsafe when the current physical context differs from the stored scenario. A plan that was valid in one season, occupancy pattern, or equipment configuration may be invalid later.

\smallskip
\noindent\textbf{Digital-twin mismatch.} Simulation is only useful when the digital twin is sufficiently calibrated. If the twin misrepresents physical dynamics, constraints, or occupant behavior, it may approve a plan that fails in deployment.

\smallskip
\noindent\textbf{Latency and delayed feedback.} Cloud planning, network congestion, or slow device responses can destabilize closed-loop behavior. A physically safe system must specify what happens when decisions arrive late.

\smallskip
\noindent\textbf{Cascading failures.} A local action can propagate across physical and digital dependencies. For example, a security lockdown may alter HVAC zones, occupant movement, emergency access, and monitoring loads.

\smallskip
\noindent\textbf{Privacy and data leakage.} IoAT agents may process sensitive occupancy, health, location, video, access-control, and operational data. Agent memory and cross-agent communication create new data-governance requirements.

\subsection{Safeguards}

The following safeguards should be built into IoAT architectures.

\begin{table}[t!]
\centering
\caption{Representative IoAT risks and safeguards.}
\label{tab:safeguards}
\begin{tabularx}{\textwidth}{p{0.30\textwidth}YY}
\toprule
\textbf{Risk} & \textbf{Safeguard} & \textbf{Operational effect} \\
\midrule
Prompt injection & Instruction hierarchy, input sanitization, tool-call validation, content provenance & Prevents untrusted text from overriding policies or device-safety constraints \\
Unsafe delegation & Least-privilege tools, capability certificates, allow-listed agent calls & Limits what each agent can do and makes delegation auditable \\
Stale memory & Scenario matching, freshness checks, validation before reuse & Ensures cached plans are reused only when context is compatible \\
Digital-twin mismatch & Calibration, uncertainty bounds, online residual monitoring & Prevents overconfidence in simulated outcomes \\
Latency & Edge fallback policies, deadlines, local safety interlocks & Maintains safe operation when cloud reasoning is delayed \\
Cascading failures & Dependency graphs, containment zones, staged execution & Limits propagation across devices and subsystems \\
Privacy leakage & Data minimization, access logging, encryption, retention limits & Protects sensitive physical and operational data \\
\bottomrule
\end{tabularx}
\end{table}

These safeguards should be enforced at multiple levels. The agentic layer should validate tool calls and maintain provenance. The edge layer should enforce device limits and fallback policies. The physical layer should maintain hard safety interlocks. The governance layer should provide audit logs, incident records, and human review for high-consequence decisions. Zero-trust principles are particularly relevant because IoAT agents should not assume that a message, tool response, device claim, or delegated action is trustworthy merely because it originates inside the network boundary \cite{nist2020zero}. For Physical AI, this layered enforcement is essential: an agent may be allowed to reason broadly, but its authority to act physically must be narrowed by capability certificates, safety envelopes, local interlocks, and escalation rules that are specific to the device, environment, and consequence class. The distinction between robustness and resilience is also important here: IoAT should not only resist disturbances within a design envelope, but also prepare for, absorb, recover from, and learn after disruptions that exceed nominal assumptions \cite{zhubasar2024resiliencerobustness}.

\subsection{Reliability Metrics}

IoAT reliability cannot be captured by task success alone. A useful evaluation suite should measure whether the user intent is satisfied, whether physical state and action constraints are respected, and how much time elapses from intent interpretation to planning, action, feedback, and recovery. It should also measure the energy and operating cost consumed by both the physical system and the agent network, the quality of replanning when deviations occur, and robustness under noisy sensors, missing data, tool failures, and communication delays. Because IoAT is a distributed form of Physical AI, evaluation must also ask whether the system preserves safe embodiment: actions should stay within device limits, respect human presence, avoid hazardous coupling among subsystems, and degrade gracefully when perception or communication fails. Security must be evaluated through resistance to prompt injection, unauthorized tool use, agent compromise, and data leakage, while human alignment should be assessed through interpretability, escalation quality, and operator trust. Such metrics should be connected to a resilience lifecycle that includes preparation, operation under attack or failure, recovery, and learning, rather than only measuring nominal performance \cite{zhu2024cyberresiliencefoundations,zhubasar2024resiliencerobustness}.

These metrics reflect the fact that IoAT is a socio-technical control system. It must be useful, efficient, safe, secure, understandable, and governable.

%% file: research_challenges.tex
The IoAT framework points to several research challenges.

\subsection{Workflow Design Under Constraints}

The first challenge is to design agent workflows that satisfy precedence, resource, logical, safety, and delay constraints. A workflow planner must know which subtasks can be parallelized, which subtasks require physical confirmation, which actions require authorization, and which device commands are incompatible. The planner must also produce specifications that are meaningful to the tactical control layer. This creates a need for formal workflow languages that are both agent-readable and control-aware, extending recent work on distributed agentic workflows toward cyber-physical execution \cite{zhu2026ioai,yangzhu2026ioaiworkflow}.

\subsection{Capability Discovery and Coalition Formation}

IoAT agents will be distributed across cloud services, organizational networks, edge devices, and local controllers. The system must discover which agents are available, what capabilities they provide, what data they can access, and what costs or risks they introduce. Coalition formation becomes a technical and economic problem: the system should select the smallest reliable set of agents that can complete the workflow under capability, locality, privacy, and incentive constraints. This challenge connects directly to incentive-compatible distributed teaming and to game-theoretic models of cooperative and noncooperative control in socio-technical systems \cite{yangzhu2026ioaiworkflow,basar2026sociotechnical}.

\subsection{Digital-Twin-Guided Planning}

Digital twins are essential for safe planning, but they introduce questions of calibration, uncertainty, and computational cost. Future work should develop methods for deciding when a digital twin is accurate enough to support a decision, how uncertainty should be communicated to the planner, and how simulation outputs should be cached for future reuse. Digital-twin agents should also be evaluated as part of a closed loop, not only as offline simulators.

\subsection{Feedback-Aware Planning and Control}

Planning and control are often separated: planners choose goals or schedules, while controllers track them. IoAT requires deeper coupling. Plans should account for physical feasibility, latency, uncertainty, and fallback behavior. Controllers should provide structured feedback that helps the agentic planner update the workflow. This calls for cross-layer methods that combine temporal planning, model predictive control, runtime verification, secure switching among control modes, and adaptive replanning \cite{miao2017movinghorizon,zhu2024cyberresiliencefoundations}.

\subsection{Physical AI Integration and Distributed Embodiment}

IoAT also raises a fundamental Physical AI question: what does embodiment mean when the intelligent system is not a single robot but a network of devices, controllers, agents, and people? In a smart building, the body of the system is distributed across locks, cameras, HVAC equipment, lighting, meters, elevators, access-control systems, mobile devices, and human interfaces. In a hospital, it may include clinical sensors, robotic carts, patient rooms, alarms, ventilation systems, and emergency workflows. In a factory, it may include robots, conveyors, programmable logic controllers, quality sensors, and maintenance systems. IoAT research must therefore develop representations of distributed embodiment that capture physical affordances, action semantics, safety envelopes, contact constraints, spatial locality, and temporal coupling across subsystems.

This challenge is not only representational. A Physical AI system must know what it can safely do in a particular environment. An agent should distinguish between issuing a reversible recommendation, changing a setpoint, moving a robot, unlocking a door, disabling an alarm, or triggering an emergency shutdown. These actions differ in reversibility, latency, uncertainty, human impact, and required authorization. IoAT needs models that make these distinctions explicit so that agentic planning can reason about physical consequence rather than treating all tools as interchangeable API calls.

\subsection{Trustworthy Agentic Execution}

IoAT actions can affect buildings, vehicles, factories, hospitals, and critical infrastructure. Trustworthy execution requires more than model accuracy. It requires constrained tool use, access control, provenance, auditability, explainability, privacy protection, and human intervention policies. The system must also distinguish between reversible and irreversible actions. A thermostat adjustment, a door unlock, a medical-device command, and an emergency shutdown should not be governed by the same approval logic. Agentic cyber-resilience frameworks suggest that attacker and defender workflows should be modeled as coupled adaptive processes, which is especially relevant when IoAT agents can both defend and operate physical systems \cite{lidzhu2025agenticcyberresilience}.

\subsection{Resilience Against Adversaries and Failures}

IoAT expands the attack surface. Adversaries may target sensors, device APIs, agent memories, prompts, digital twins, or A2A communication channels. Robust architectures should integrate cyber defense, anomaly detection, redundancy, diversity, and containment. A promising direction is to use specialized security agents that continuously test workflows, inspect tool calls, and validate physical actions before execution. Game-theoretic and resilience-oriented models are useful here because they treat adversaries as strategic, adaptive agents rather than as fixed disturbances \cite{zhu2024cyberresiliencefoundations,zhubasar2024resiliencerobustness,zhubasar2024sociotechnical}.

\subsection{Benchmarks and Case Studies}

The field needs benchmark environments for evaluating IoAT systems. Smart buildings provide a natural starting point because they include sensing, actuation, energy objectives, comfort constraints, security policies, and human occupancy. Other benchmark domains include hospital IoT, industrial automation, connected transportation, smart grids, and autonomous laboratories. Benchmarks should include both concentrated embodiment, such as robots or vehicles, and distributed embodiment, such as buildings or infrastructure networks, so that the connection between IoAT and Physical AI can be evaluated across different physical forms. They should measure not only task success but also physical safety, latency, cost, robustness, operator workload, privacy, and recovery from disruptions.

These challenges suggest that IoAT is not a single application but a research program at the intersection of multi-agent systems, IoT, cyber-physical systems, Physical AI, control theory, cybersecurity, human-agent interaction, and AI governance.